\documentclass[applsci,article,accept,moreauthors,pdftex]{Definitions/mdpi} 

\firstpage{1} 
\pubvolume{1}
\issuenum{1}
\articlenumber{0}
\pubyear{2021}
\copyrightyear{2021}
\externaleditor{Academic Editor: Chris G. Tzanis}
\datereceived{29 April 2021} 
\dateaccepted{11 June 2021} 
\datepublished{} 
\hreflink{https://doi.org/} % Please use \linebreak if need to do line break
\usepackage[per-mode=fraction]{siunitx}
\usepackage[inkscapearea=page]{svg}
\usepackage{adjustbox}
\usepackage{bm}
\usepackage{subfigure}
\usepackage{graphicx,import}
\usepackage{arydshln}

\usepackage{subfigure}
\makeatletter
\renewcommand{\@thesubfigure}{\normalsize(\textbf{\alph{subfigure}})}
\makeatother

\graphicspath{ {./figures/} }

\Title{Radar Voxel Fusion for 3D Object Detection}

\TitleCitation{Radar Voxel Fusion for 3D Object Detection}

\Author{Felix Nobis $^{1,}$*\orcidA{}, Ehsan Shafiei $^{1}$, Phillip Karle $^{1}$, Johannes Betz $^{2}$\orcidB{} and Markus Lienkamp $^{1}$}
\AuthorNames{Felix Nobis, Ehsan Shafiei, Phillip Karle, Johannes Betz and Markus Lienkamp}

\AuthorCitation{Nobis, F.; Shafiei, E.; Karle P.; Betz, J.; Lienkamp, M.}
\address{%
$^{1}$ \quad Institute of Automotive Technology, Technical University of Munich, 85748 Garching, Germany; nobis@ftm.mw.tum.de (F.N.); ehsan.shafiei@tum.de (E.S.); karle@ftm.mw.tum.de (P.K.); lienkamp@ftm.mw.tum.de (M.L.) \\
$^{2}$ \quad mLab:Real-Time and Embedded Systems Lab, University of Pennsylvania, PA 19104, USA; joebetz@seas.upenn.edu (J.B.)
}
\corres{Correspondence: nobis@ftm.mw.tum.de}

\abstract{Automotive traffic scenes are complex due to the variety of possible scenarios, objects, and weather conditions  that need to be handled. In contrast to more constrained environments, such as automated underground trains, automotive perception systems cannot be tailored to a narrow field of specific tasks but must handle an ever-changing environment with unforeseen events. As currently no single sensor is able to reliably perceive all relevant activity in the surroundings, sensor data fusion is applied to perceive as much information as possible. Data fusion of different sensors and sensor modalities on a low abstraction level enables the compensation of sensor weaknesses and misdetections among the sensors before the information-rich sensor data are compressed and thereby information is lost after a sensor-individual object detection. This paper develops a low-level sensor fusion network for 3D object detection, which fuses lidar, camera, and radar data. The fusion network is trained and evaluated on the nuScenes data set. On the test set, fusion of radar data increases the resulting AP (Average Precision) detection score by about \SI{5.1}{\percent} in comparison to the baseline lidar network. The radar sensor fusion proves especially beneficial in inclement conditions such as rain and night scenes. Fusing additional camera data contributes positively only in conjunction with the radar fusion, which shows that interdependencies of the sensors are important for the detection result. Additionally, the paper proposes a novel loss to  handle the discontinuity of a simple yaw representation for object detection. Our updated loss increases the detection and orientation estimation performance for all sensor input configurations. The code for this research has been made available on GitHub.}


\begin{document}
%%%%%%%%%%%%%%%%%%%%%%%%%%%%%%%%%%%%%%%%%%
\section{Introduction}

In the current state of the art, researchers focus on 3D object detection in the field of perception. Three-dimensional object detection is most reliably performed with lidar sensor data~\cite{Caesar.2019, Geiger.2013,Shi.31.12.2019} as its higher resolution---when compared to radar sensors---and direct depth measurement---when compared to camera sensors---provide the most relevant features for object detection algorithms. However, for redundancy and safety reasons in autonomous driving applications, additional sensor modalities are required because lidar sensors  cannot detect all relevant objects at all times. Cameras are well-understood, cheap and reliable sensors for applications such as traffic-sign recognition. Despite their high resolution, their capabilities for 3D perception are limited as only 2D information is provided by the sensor. Furthermore, the sensor data quality deteriorates strongly in bad weather conditions such as snow or heavy rain. Radar sensors are least affected by inclement weather, e.g., fog, and are therefore a vital asset to make autonomous driving more reliable. However, due to their low resolution and clutter noise for static vehicles, current radar sensors cannot perform general object detection without the addition of further modalities. This work therefore combines the advantages of camera, lidar, and radar sensor modalities to produce an improved detection result. 

Several strategies exist to fuse the information of different sensors. These systems can be categorized as {early fusion} if all input data are first combined and then processed, or {late fusion} if all data is first processed independently and the output of the data-specific algorithms are fused after the processing. Partly independent and joint processing is called middle or {feature fusion}.

Late fusion schemes based on a Bayes filter, e.g., the Unscented Kalman Filter (UKF)~\cite{Julier.1997}, in combination with a matching algorithm for object tracking, are the current state of the art, due to their simplicity and their effectiveness during operation in constrained environments and good weather.

Early and feature fusion networks possess the advantage of using all available sensor information at once and are therefore able to learn from interdependencies of the sensor data and compensate imperfect sensor data for a robust detection result similar to gradient boosting~\cite{Friedman.2001}.

This paper presents an approach to fuse the sensors in an early fusion scheme.  Similar to Wang et al.~\cite{Wang.1019202011132020}, we color the lidar point cloud with camera RGB information. These colored lidar points are then fused with the radar points and their  radar cross-section (RCS) and velocity features. The network processes the points jointly in a voxel structure and outputs the predicted bounding boxes. The paper evaluates several parameterizations and presents the {RadarVoxelFusionNet {(}RVF-Net{),}} which proved most reliable in our studies.  

The contribution of the paper is threefold:

\begin{itemize}
    \item The paper develops an early fusion network for radar, lidar, and camera data for 3D object detection. The network outperforms the lidar baseline and a Kalman Filter late fusion approach.
    \item The paper provides a novel loss function to replace the simple discontinuous yaw parameterization during network training.
    \item The code for this research has been released to the public to make it adaptable to further use cases.
\end{itemize}

Section \ref{sec:related_work} discusses related work for object detection and sensor fusion networks. The proposed model is described in Section \ref{sec:methodology}. The results are shown in Section \ref{sec:evaluation} and discussed in Section \ref{sec:discussion}. Section \ref{sec:conclusions} presents our conclusions from the work.

%%%%%%%%%%%%%%%%%%%%%%%%%%%%%%%%%%%%%%%%%%
 \section{Related Work}
 \label{sec:related_work}
\textls[-15]{Firstly, this section gives a short overview of the state of the art of lidar object detection for autonomous driving. Secondly, a more detailed review of fusion methods for object detection is given. We refer to~\cite{Nobis.2021} for a more detailed overview of radar object detection methods.}

\subsection{3D Lidar Object Detection}

The seminal work of Qi et al.~\cite{Qi.2017} introduces a method to directly process sparse, irregular point cloud data with neural networks for semantic segmentation tasks. Their continued work~\cite{Qi.20180413} uses a similar backbone to perform 3D object detection from point cloud frustums. Their so-called pointnet backbone has been adapted in numerous works to advance lidar object detection.

VoxelNet~\cite{Zhou.17.11.2017} processes lidar points in a voxel grid structure. The network aggregates a feature for each voxel from the associated points. These voxel grid cells are processed in a convolutional fashion to generate object detection results with an anchor-based region proposal network (RPN)~\cite{Ren.2015}.

The network that achieves the highest object detection score~\cite{Shi.31.12.2019} on the KITTI 3D benchmark~\cite{Geiger.2013} uses both a voxel-based and pointnet-based processing to create their detection results. The processing of the voxel data is performed with submanifold sparse convolutions as introduced in~\cite{Graham.05.06.2017, Graham.28.11.2017}. The advantage of these sparse implementation of convolutions lies in the fact that they do not process empty parts of the grid that contain no information. This is especially advantageous for point cloud processing, as most of the 3D space does not contain any sensor returns. 
The network that achieves the highest object detection score on the nuScenes data set~\cite{Caesar.2019} is a lidar-only approach as well~\cite{Yin.19.06.2020}. Similarly, it uses a sparse VoxelNet backbone with a second stage for bounding box refinement.

\subsection{2D Sensor Fusion for Object Detection}
 
This section reviews 2D fusion methods. The focus is on methods that fuse radar data as part of the input data.

Chadwick~\cite{Chadwick.2019} is the first to use a neural network to fuse low level radar and camera for 2D object detection. The network fuses the data on a feature level after projecting radar data to the 2D image plane. The object detection scores of the fusion are higher than the ones of a camera-only network, especially for distant objects.  

CRF-Net~\cite{Nobis.2019} develops a similar fusion approach. As an automotive radar does not measure any height information, the network assumes an extended height of the radar returns to account for the uncertainty in the radar returns origin. The approach shows a slight increase in object detection performance both on a private and the public nuScenes data set~\cite{Caesar.2019}. The paper shows further potential for the fusion scheme once less noisy radar data are available.  

YOdar \cite{Kowol.07.10.2020} uses a similar projection fusion method. The approach creates two detection probabilities of separate radar and image processing pipelines and generates their final detection output by gradient boosting.

\subsection{3D Sensor Fusion for Object Detection}

This section reviews 3D fusion methods. The focus is on methods that fuse radar data as part of the input data.

\subsubsection{Camera Radar Fusion}

For 3D object detection,~the authors of \cite{Kim.2020} propose GRIF-Net to fuse radar and camera data. After individual processing, the feature fusion is performed by a gated region of interest fusion (GRIF). In contrast to concatenation or addition as the fusion operation, the weight for each sensor in the fusion is learned in the GRIF module. The camera and radar fusion method outperforms the radar baseline by a great margin on the nuScenes data set.

The CenterFusion architecture~\cite{Nabati.10.11.2020} first detects objects in the 3D space via image-based object detection. Radar points inside a frustum around these detections are fused by concatenation to the image features. The radar features are extended to pillars similar to~\cite{Nobis.2019} in the 2D case. The object detection head operates on these joint features to refine the detection accuracy. The mean Average Precision (mAP) score of the detection output increases by 4\% for the camera radar fusion compared to their baseline on the nuScenes validation data set.

While the methods above operate with point cloud-based input data, Lim~\cite{Lim.2019} fuses azimuth range images and camera images. The camera data are projected to a bird’s-eye view (BEV) with an Inverse Projection Mapping (IPM). The individually processed branches are concatenated to generate the object detection results. The fusion approach achieves a higher detection score than the individual modalities. The IPM limits the detection range to close objects and an assumed flat road surface. 

Kim~\cite{Kim.2020b} similarly fuses radar azimuth-range images with camera images. The data are fused after initial individual processing, and the detection output is generated adopting the detection head of~\cite{Ku.2018}. The fusion approach outperforms both their image and radar baselines on their private data set. Their RPN uses a distance threshold in contrast to standard Intersection over Union (IoU) matching for anchor association. The paper argues that the IoU metric prefers to associate distant bounding boxes over closer bounding boxes under certain conditions. Using a distance threshold instead increases the resulting AP by 4--5 points over the IoU threshold matching.

The overall detection accuracy of camera radar fusion networks is significantly lower than that of lidar-based detection methods.

\subsubsection{Lidar Camera Fusion}

MV3D~\cite{Chen.2016} projects lidar data both to a BEV perspective and the camera perspective. The lidar representations are fused with the camera input after some initial processing in a feature fusion scheme. 

AVOD~\cite{Ku.2018} uses a BEV projection of the lidar data and camera data as their input data. The detection results are calculated with an anchor grid and an RPN as a detection head. 

PointPainting~\cite{Vora.22.11.2019} first calculates a semantic segmentation mask for an input image. The detected classes are then projected onto the lidar point cloud via a color-coding for the different classes. The work expands several lidar 3D object detection networks and shows that enriching the lidar data with class information augments the detection score.

\subsubsection{Lidar Radar Fusion}

\textls[-15]{RadarNet~\cite{Yang.28.07.2020} fuses radar and lidar point clouds for object detection. The point clouds are transformed into a grid representation and then concatenated. After this feature fusion, the data are processed jointly to propose an object detection. An additional late fusion of radar features is performed to predict a velocity estimate separate to the object detection task.}

\subsubsection{Lidar Radar Camera Fusion}

Wang~\cite{Wang.1019202011132020} projects RGB values of camera images directly onto the lidar point cloud. This early fusion camera-lidar point cloud is used to create object detection outputs in a pointnet architecture. Parallel to the object detection, the radar point cloud is processed to predict velocity estimates of the input point cloud. The velocity estimates are then associated with the final detection output. The paper experimented with concatenating different amounts of past data sweeps for the radar network. Concatenating six consecutive time steps of the radar data for a single processing shows the best results in their study. The addition of the radar data increases their baseline detection score slightly on the public nuScenes data set.

%%%%%%%%%%%%%%%%%%%%%%%%%%%%%%%%%%%%%%%%%%
\section{Methodology}
\label{sec:methodology}

In the following, we list the main conclusions from the state of the art for our work:
\begin{itemize}[leftmargin=*,labelsep=5.8mm]
\item \emph{Input representation}: The input representation of the sensor data dictates which subsequent processing techniques can be applied. Pointnet-based methods are beneficial when dealing with sparse unordered point cloud data. For more dense{---but still sparse---}point clouds, such as the fusion of several lidar or radar sensors, {sparse} voxel grid structures achieve more favorable results in the object detection literature.  Therefore, we adopt a voxel-based input structure for our data. As many of the voxels remain empty in the 3D grid, we apply sparse convolutional operations~\cite{Graham.05.06.2017} for greater efficiency.

\item \emph{Distance Threshold}: Anchor-based detection heads predominately use an IoU-based matching algorithm to identify positive anchors. However, Kim~\cite{Kim.2020b} has shown that this choice might lead to association of distant anchors for certain bounding box configurations. We argue that both IoU- and distance-based matching thresholds should be considered to facilitate the learning process. The distance-based threshold alone might not be a good metric when considering rotated bounding boxes with a small overlapping area. Our network therefore considers both thresholds to match the anchor boxes. 

\item \emph{Fusion Level}: The data from different sensors and modalities can be fused at different abstraction levels. Recently, a rising number of papers perform early or feature fusion to be able to facilitate all available data for object detection simultaneously. Nonetheless, the state of the art in object detection is still achieved by considering only lidar data. Due to its resolution and precision advantage from a hardware perspective, software processing methods cannot compensate for the missing information in the input data of the additional sensors. Still, there are use cases where the lidar sensor alone is not sufficient. Inclement weather, such as fog, decreases the lidar and camera data quality~\cite{Daniel.2017} significantly. The radar data, however, is only slightly affected by the change in environmental conditions. Furthermore, interference effects of different lidar modules might decrease the detection performance under certain conditions~\cite{Hebel.2018, Kim.2015}. A drawback of early fusion algorithms is that {temporal synchronized data recording} for all sensors needs to be available. However, none of the publicly available data sets provide such data for all three sensor modalities.~The authors of \cite{Nobis.2021} discuss the publicly available data quality for radar sensors in more detail. Despite the lack of synchronized data, this study uses an early fusion scheme, as in similar works, {spatio-temporal} synchronization errors are treated as noise and compensated during the learning process of the fusion network. In contrast to recent papers, where some initial processing is applied before fusing the data, we present a direct early fusion to enable the network to learn optimal combined features for the input data. {The early fusion can make use of the complementary sensor information provided by radar, camera and lidar sensors---before any data compression by sensor-individual processing is performed.}
\end{itemize}

\subsection{Input Data}

The input data to the network consists of the lidar data with its three spatial coordinates $x$, $y$, $z$, and intensity value $i$. Similar to~\cite{Wang.1019202011132020}, colorization from projected camera images is added to the lidar data with $r$, $g$, $b$ features. Additionally, the radar data contributes its spatial coordinates, intensity value $RCS$---and the radial velocity with its Cartesian components $v_x$ and $v_y$. Local offsets for the points in the voxels $dx$, $dy$, $dz$ complete the input space. The raw data are fused and processed jointly by the network itself. Due to the early fusion of the input data, any lidar network can easily be adapted to our fusion approach by adjusting the input dimensions.

\subsection{Network Architecture}
\label{sec:network_arch}

This paper proposes the RadarVoxelFusionNet (RVF-Net) whose architecture is based on VoxelNet~\cite{Zhou.17.11.2017} due to its empirically proven performance and straightforward network architecture. While other architectures in the state of the art provide higher detection scores, the application to a non-overengineered network from the literature is preferable for investigating the effect of a new data fusion method. Recently, A. Ng~\cite{Ng.2021} proposed a shift from model-centric to data-centric approaches for machine learning development.  

\begin{figure}[H]
    
    \includegraphics[width=13.4cm]{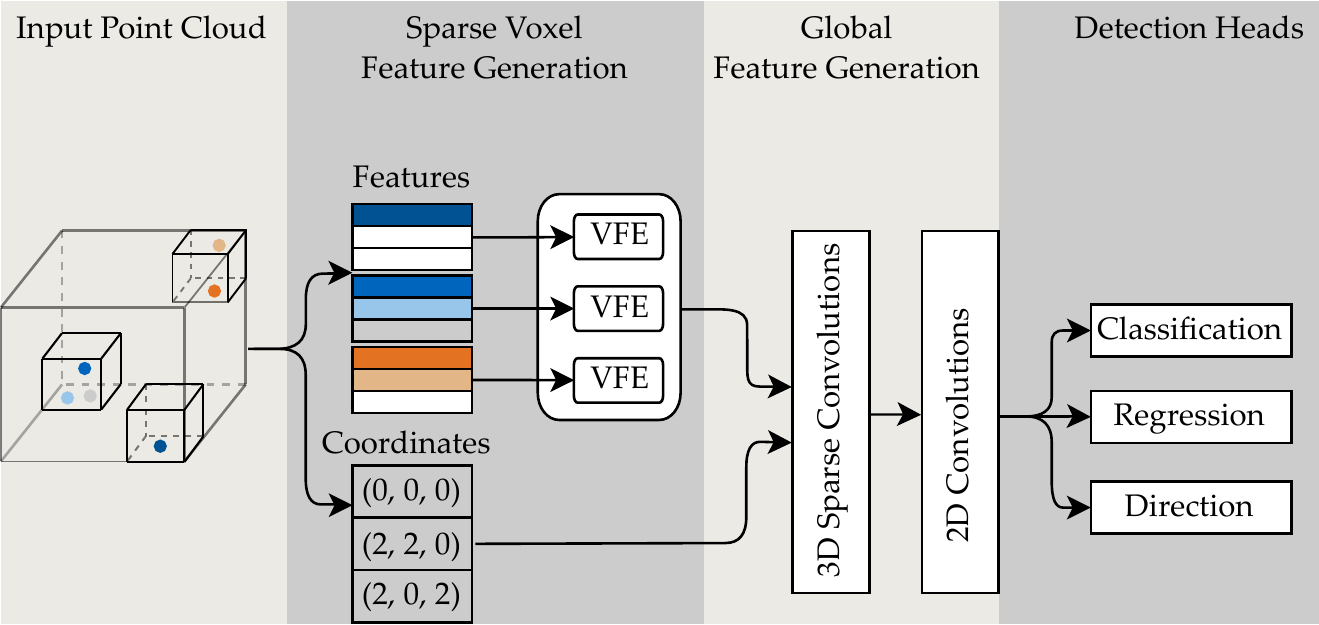}
    \caption{Network architecture of the proposed RVF-Net.}
    \label{fig:network_architecture}
\end{figure}

An overview of the network architecture is shown in Figure \ref{fig:network_architecture}. The input point cloud is partitioned into a 3D voxel grid. Non-empty voxel cells are used as the input data to the network. The data are split into the features of the input points and the corresponding coordinates. The input features are processed by voxel feature encoding (VFE) layers composed of fully connected and max-pooling operations for the points inside each voxel. The pooling is used to aggregate one single feature per voxel. In the global feature generation, the voxel features are processed by sparse 3D submanifold convolutions to efficiently handle the sparse voxel grid input. The $z$ dimension is merged with the feature dimension to create a sparse feature tensor in the form of a 2D grid. The sparse tensor is converted to a dense 2D grid and processed with standard 2D convolutions to generate features in a BEV representation. These features are the basis for the detection output heads.

The detection head consists of three parts: The classification head, which outputs a class score for each anchor box; the regression head with seven regression values for the bounding box position ($x$, $y$, $z$), dimensions ($w$, $l$, $h$) and the yaw angle $e_{\theta}$; the direction head, which outputs a complementary classification value for the yaw angle estimation $c_{dir}$. For more details on the network architecture, we refer to the work of~\cite{Zhou.17.11.2017} and our open source implementation. The next section focuses on our proposed yaw loss, which is conceptually different from the original VoxelNet implementation.

\subsection{Yaw Loss Parameterization}
\label{sec:loss}

\begin{figure}[H]
\centering
\includegraphics{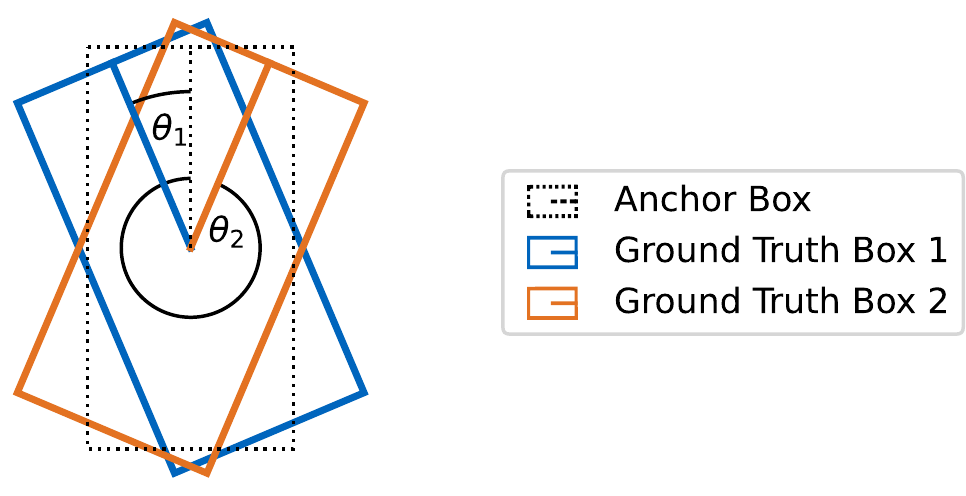}
\begin{equation*}
\begin{split}
        \theta_2  &\gg \theta_1 \\
      L(\theta_2) &\gg L(\theta_1)
\end{split}
\end{equation*}
\caption{Vehicle bounding boxes are visualized in a BEV. The heading of the vehicles is visualized with a line from the middle of the bounding box to the front. The relative angular deviations from the orange and blue ground truth boxes to the anchor box are equal. However, the resulting loss value of the orange bounding box is significantly higher than that of the blue one.} 
\label{fig:yaw_diff}
\end{figure}

While the original VoxelNet paper uses a simple yaw regression, we use a more complex parameterization to facilitate the learning process. Zhou~\cite{Zhou.17.12.2018} argues that a simple yaw representation is disadvantageous, as the optimizer needs to regress a smooth function over a discontinuity, e.g., from $\SI[number-math-rm=\mathnormal]{-\pi}{\radian}$ to +$\SI[number-math-rm=\mathnormal]{\pi}{\radian}$. Furthermore, the loss value for small positive angle differences is much lower than that of greater positive angle differences, while the absolute angle difference from the anchor orientation might be the same. \mbox{Figure \ref{fig:yaw_diff}} visualizes this problem of an exemplary simple yaw regression.

To account for this problem, the network estimates the yaw angle with a combination of a classification and a regression head. The classifier is inherently designed to deal with a discontinuous domain, enabling the regression of a continuous target. The regression head regresses the actual angle difference in the interval $[-\pi, \pi)$ with a smooth sine function, which is continuous even at the limits of the interval. The regression output of the yaw angle of a bounding box is
\begin{equation}
\begin{split}
\theta_{d} &= \theta_{GT} - \theta_{A} \\
 e_{\theta} &= sin(\theta_{d}),
 \end{split}
\end{equation}
where $\theta_{GT}$ is the ground truth box yaw angle and $\theta_{A}$ is the associated anchor box \mbox{yaw angle. }

The classification head determines whether or not the angle difference between the predicted bounding box and the associated anchor lies inside or outside of the interval $[-\pi/2, \pi/2)$.  The classification value of the yaw is modeled as
\begin{equation}
c_{dir} =   
\begin{cases}
    1,& \text{if } -\frac{\pi}{2} \leq  (\theta_{d}+ \pi) \bmod 2 \pi - \pi < \frac{\pi}{2} \\
    0,              & \text{otherwise}
\end{cases}
.
\end{equation}

As seen above, the directional bin classification head splits the angle space into two equally spaced regions. The network uses two anchor angle parameterizations at $0$ and $\pi/2$. A vehicle driving towards the sensor vehicle matches with the anchor at $\SI{0}{\radian}$. A vehicle driving in front of the vehicle would match with the same anchor. The angle classification head intuitively distinguishes between these cases. Therefore, there is no need to compute additional anchors at $\pi$ and $-\pi/2$.

\begin{figure}[H]
\centering
\subfigure[]{\includegraphics{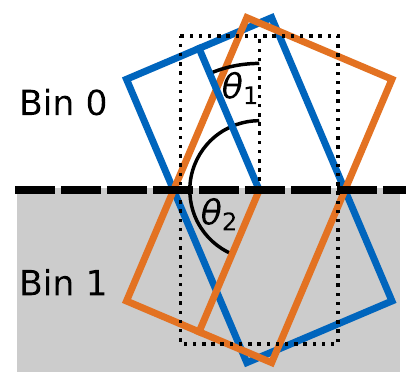}}
\subfigure[]{\includegraphics{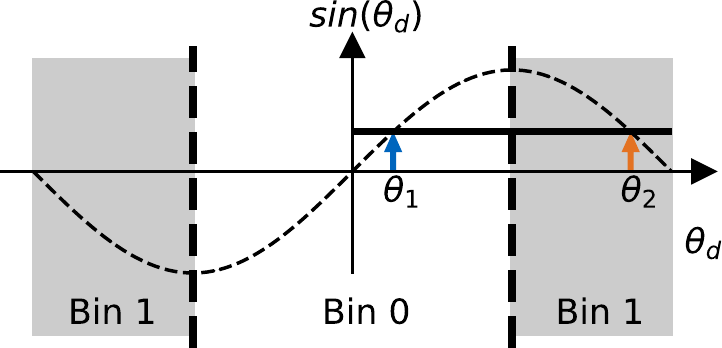}}
\caption{Visualization of our yaw loss. ({\bf a}) Bin classification. ({\bf b}) Sine regression. The bounding boxes in ({\bf a}) are not distinguishable by the sine loss. The bin classification distinguishes these bounding boxes as visualized by the bold dotted line, which splits the angular space in two parts. }
\label{fig:yaw_loss}
\end{figure}

Due to the subdivision of the angular space by the classification head, the yaw regression needs to regress smaller angle differences, which leads to a fast learning progress. A simple yaw regression would instead need to learn a rotation of 180 degrees to match the ground truth bounding box. It has been shown that high regression values and discontinuities negatively impact the network performance~\cite{Zhou.17.12.2018}. The regression and classification losses used to estimate the yaw angle are visualized in Figure \ref{fig:yaw_loss}. 

The SECOND architecture~\cite{Yan.2018} introduces a sine loss as well. Their subdivision of the positive and negative half-space, however, comes with the drawback that both bounding box configurations shown in Figure \ref{fig:yaw_loss} would result in the same regression and classification loss values. Our loss is able to distinguish these bounding box configurations.

As the training does not learn the angle parameter directly, the regression difference is added to the anchor angle under consideration of the classification interval output to get the final value of the yaw angle during inference. 

\subsection{Data Augmentation}

Data augmentation techniques~\cite{Shorten.2019} manipulate the input features of a machine learning method to create a greater variance in the data set. Popular augmentation methods translate or rotate the input data to generate  \emph{new} input data from the existing data set. 

More complex data augmentation techniques include the use of General Adversarial Networks~\cite{Sandfort.2019} to generate artificial data frames in the style of the existing data. Complex data augmentation schemes are beneficial for small data sets. The used nuScenes data set comprises about 34,000 labeled frames. Due to the relatively large data set, we limit the use of data augmentation to rotation, translation, and scaling of the input point cloud.

\subsection{Inclement Weather}

We expect the fusion of different sensor modalities to be most beneficial in inclement weather, which deteriorates the quality of the output of lidar and camera sensors. We analyze the nuScenes data set for frames captured in such environment conditions. At the same time, we make sure that enough input data, in conjunction with data augmentation, are available for the selected environment conditions to realize a good generalization for the trained networks. We filter the official nuScenes training and validation sets for samples recorded in rain or night conditions. Further subsampling for challenging conditions such as fog is not possible for the currently available data sets. The amount of samples for each split is shown in Table \ref{tab:training_splits}. We expect the lidar quality to deteriorate in the rain scenes, whereas the camera quality should deteriorate in both rain and night scenes. The radar detection quality should be unaffected by the environment conditions. 

\begin{specialtable}[H]
    
    \caption{Training and validation splits for different environment conditions. The table only considers samples in which at least one car is present in the field of view of the front camera.}
    \label{tab:training_splits}
\setlength{\cellWidtha}{\columnwidth/3-2\tabcolsep+0.0in}
\setlength{\cellWidthb}{\columnwidth/3-2\tabcolsep+0.0in}
\setlength{\cellWidthc}{\columnwidth/3-2\tabcolsep+0.0in}
\scalebox{1}[1]{\begin{tabularx}{\columnwidth}{>{\PreserveBackslash\centering}m{\cellWidtha}>{\PreserveBackslash\centering}m{\cellWidthb}>{\PreserveBackslash\centering}m{\cellWidthc}}
    \toprule
         \textbf{Data Set Split} & \textbf{Training Samples} & \textbf{Validation Samples} \\
         \midrule
          nuScenes & 19659 & 4278 \\
          Rain & 2289 & 415 \\
          Night & 4460 & 788 \\
          \bottomrule
    \end{tabularx}
    }
\end{specialtable}

\subsection{Distance Threshold}

Similar to~\cite{Kim.2020b}, we argue that an IoU-based threshold is not the optimal choice for 3D object detection. We use both an IoU-based and a distance-based threshold to distinguish between the positive, negative, and ignore bounding box anchors. For our proposed network, the positive IoU-threshold is empirically set to \SI{35}{\percent} and the negative threshold is set to \SI{30}{\percent}. The distance threshold is set to \SI{0.5}{\meter}.

\subsection{Simulated Depth Camera}
To simplify the software fusion scheme and to lower the cost of the sensor setup, lidar and camera sensor could be replaced by a depth or stereo camera setup. Even though the detection performance of stereo vision does not match the one of lidar, recent developments show promising progress in this field~\cite{You.20190825}. The relative accuracy of stereo methods is higher for close range objects, where high accuracy is of greater importance for the planning of the driving task. The nuScenes data set was chosen for evaluation since it is the only feasible public data set that contains labeled point cloud radar data. However, stereo camera data are not included in the nuScenes data set, which we use for evaluation. 

In comparison to lidar data, stereo camera data are more dense and contain the color of objects in its data. To simulate a stereo camera, we use the IP-Basic algorithm~\cite{Ku.31.01.2018} to approximate a denser depth image from the sparser lidar point cloud. {The IP-Basic algorithm estimates additional depth measurements from lidar pixels, so that additional camera data can be used for the detection. The depth of these estimated pixels is less accurate than that of the lidar sensor, which is in compliance with the fact that stereo camera depth estimation is also more error-prone than that of lidar}~\cite{Chen.2010, Fan.2020}.

Our detection pipeline looks for objects in the surroundings of up to \SI{50}{\meter} from the ego vehicle so that the stereo camera simulation by the lidar is justified {as production stereo cameras can provide reasonable accuracy in this sensor range}~\cite{Instruments.2016, Wang.2018}. An alternative approach would be to learn the depth of the monocular camera images directly. An additional study~\cite{Nobis.92020209232020} showed that the state of the art algorithms in this field~\cite{Zhao.2020} are not robust enough to create an accurate depth estimation for the whole scene for a subsequent fusion. Although the visual impression of monocular depth images seems promising, the disparity measurement of stereo cameras results in a better depth estimation.

\subsection{Sensor Fusion}
By simulating depth information for the camera, we can investigate the influence of four different sensors for the overall detection score: radar, camera, simulated depth camera, and lidar. In addition to the different sensors, consecutive time steps of radar and lidar sensors are concatenated to increase the data density. While the nuScenes data set allows to concatenate up to 10 lidar sweeps on the official score board, we limit our network to use the past 3 radar and lidar sweep data. While using more sweeps may be beneficial for the overall detection score through the higher data density for static objects, more sweeps add significant inaccuracies for the position estimate of moving vehicles, which are of greater interest for a practical use case.

As discussed in our main conclusions from the state of the art in Section \ref{sec:methodology}, we fuse the different sensor modalities in an early fusion scheme. In particular, we fuse lidar and camera data by projecting the lidar data into the image space, where the lidar points serve as a mask to associate the color of the camera image with the 3D points.

To implement the simulated depth camera, we first apply the IP-Basic algorithm to the lidar input point cloud to approximate the depth of the neighborhood area of the lidar points to generate a more dense point cloud. The second step is the same as in the lidar and camera fusion, where the newly created point cloud serves as a mask to create the dense depth color image.  

The radar, lidar, and simulated depth camera data all originate from a continuous 3D space. The data are then fused together in a discrete voxel representation before they are processed with the network presented in Section \ref{sec:network_arch}. The first layers of the network compress the input data to discrete voxel features. The maximum number of points per voxel is limited to 40 for computational efficiency. As the radar data are much sparser than lidar data, it is preferred in the otherwise random downsampling process to make sure that the radar data contributes to the fusion result and its data density is not further reduced.

After the initial fusion step, the data are processed in the RadarVoxelFusionNet in the same fashion, independent of which data type was used. This modularity is used to compare the detection result of different sensor configurations.

\subsection{Training}
The network is trained with an input voxel size of \SI{0.2}{\meter} for the dimensions parallel to the ground. The voxel size in height direction is \SI{0.4}{\meter}. 

Similar to the nuScenes split, we limit the sensor detection and evaluation range to \SI{50}{\meter} in front of the vehicle and further to \SI{20}{\meter} on either side to cover the principal area of interest for driving. The sensor fusion is performed for the front camera, front radar, and the lidar sensor of the nuScenes data set. 

The classification outputs are learned via a binary cross entropy loss. The regression values are learned via a smooth L1 loss~\cite{Girshick.2015}. The training is performed on the official nuScenes split. We further filter for samples that include at least one vehicle in the sensor area to save training resources for samples where no object of interest is present. Training and evaluation are performed for the nuScenes car class. Each network is trained on an NVIDIA Titan Xp graphics card for 50 epochs or until overfitting can be deduced from the validation loss curves.

%%%%%%%%%%%%%%%%%%%%%%%%%%%%%%%%%%%%%%%%%%
\section{Results}
\label{sec:evaluation}

The model performance is evaluated with the average precision (AP) metric as defined by the nuScenes object detection challenge~\cite{Caesar.2019}. Our baseline is a VoxelNet-style network with lidar data as the input source. All networks are trained with our novel yaw loss and training strategies, as described in Section \ref{sec:methodology}.

\subsection{Sensor Fusion}
\label{sec:sensor_fusion}

Table \ref{table:results_fusion} shows the results of the proposed model with different input sensor data. {The networks have been trained several times to rule out that the different AP scores are caused by random effects.} The lidar baseline outperforms the radar baseline by a great margin. This is expected as the data density and accuracy of the lidar input data are higher than that of the radar data. 

The fusion of camera RGB and lidar data does not result in an increased detection accuracy for the proposed network. We assume that this is due to the increased complexity that the additional image data brings into the optimization process. At the same time, the additional color feature does not distinguish vehicles from the background, as the same colors are also widely found in the environment.

\begin{figure}[H]
    \centering
    \frame{\includegraphics[width=3cm]{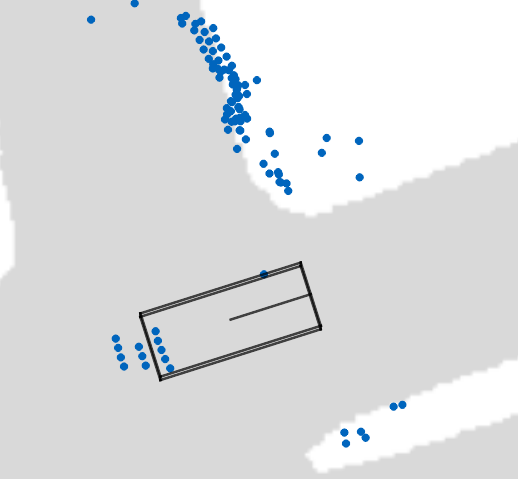}}
    \frame{\includegraphics[width=5cm]{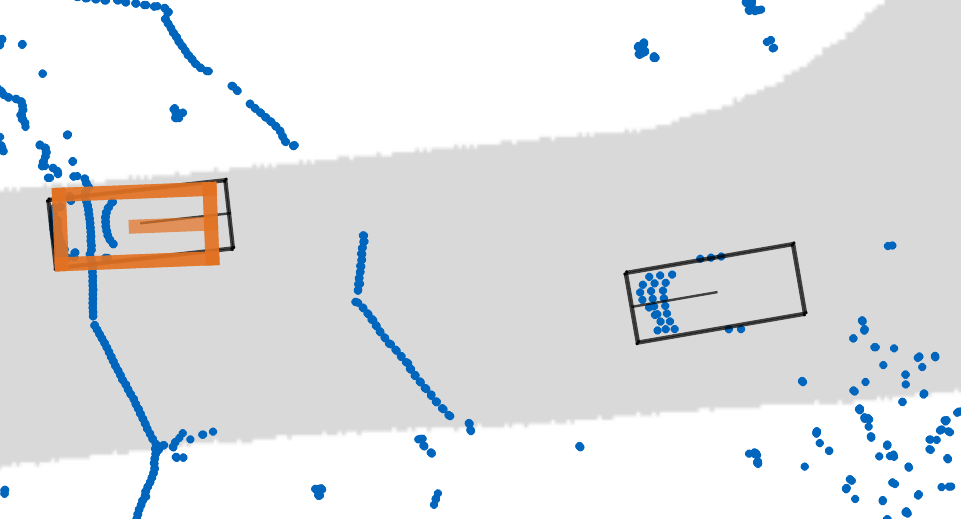}}
    \frame{\includegraphics[width=5cm]{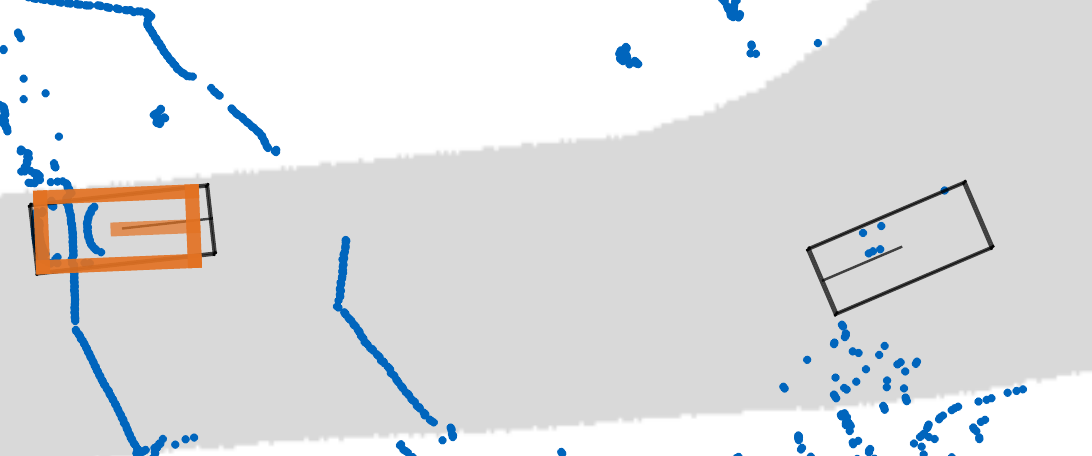}}
    \subfigure[]{\frame{\includegraphics[width=3cm]{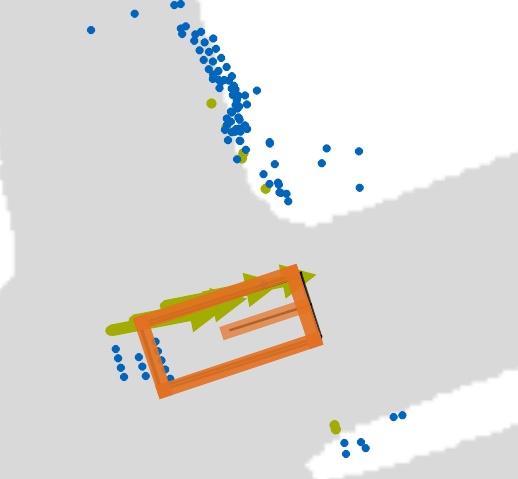}}}
    \subfigure[]{\frame{\includegraphics[width=5cm]{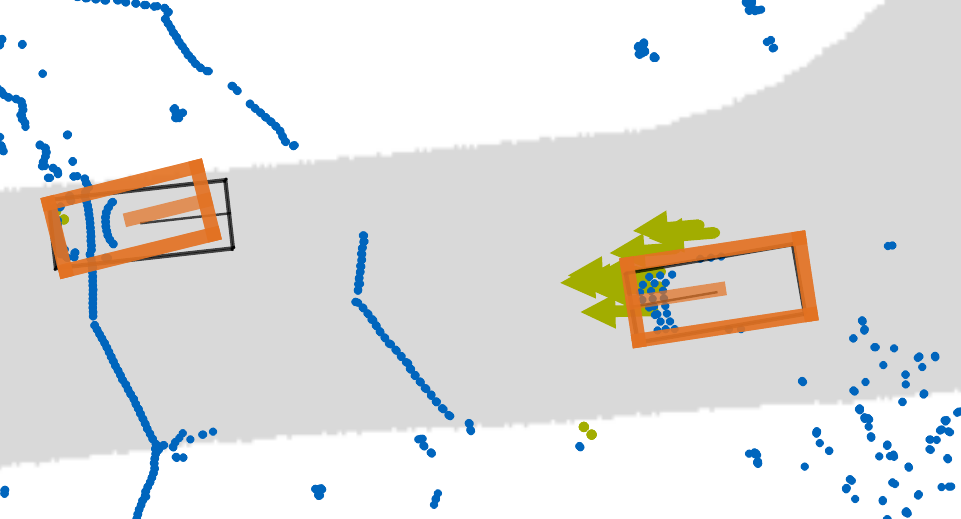}}}
    \subfigure[]{\frame{\includegraphics[width=5cm]{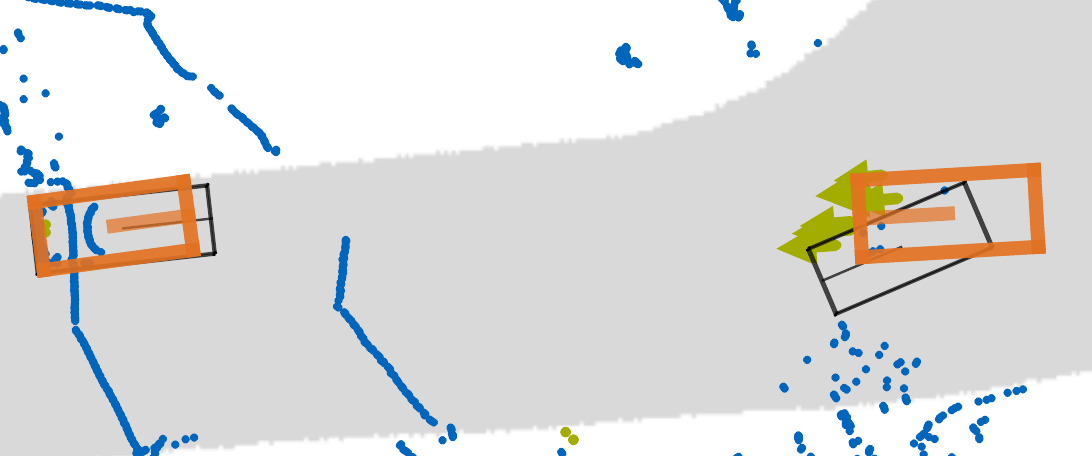}}}
    \caption{BEV of the detection results: Lidar and RGB fusion in the top row. RVF-Net fusion in the bottom row. Detected bounding boxes in orange. Ground truth bounding boxes in black. Lidar point cloud in blue. Radar point cloud and measured velocity in green. The street is shown in gray. ({\bf a})~Only RVF-Net is able to detect the vehicle from the point cloud. ({\bf b}) RVF-Net detects both bounding boxes. ({\bf c}) RVF-Net detects both bounding boxes. However, the detected box on the right has a high translation and rotation error towards the ground truth bounding box.}
    \label{fig:det_vis}
\end{figure}

The early fusion of radar and lidar data increases the network performance against the baseline. The fusion of all three modalities increases the detection performance by a greater margin for most of the evaluated data sets. Only for night scenes, where the camera data deteriorates most, does the fusion of lidar and radar outperform the RVF-Net. Example detection results in the BEV perspective from the lidar, RGB input, and the RVF-Net input are compared in Figure \ref{fig:det_vis}.

\begin{specialtable}[H] 
%\tablesize{\scriptsize}
\caption{AP scores for different environment (data) and network configurations on the respective validation data set.}
\label{table:results_fusion}
\setlength{\cellWidtha}{\columnwidth/5-2\tabcolsep+0.8in}
\setlength{\cellWidthb}{\columnwidth/5-2\tabcolsep-0.2in}
\setlength{\cellWidthc}{\columnwidth/5-2\tabcolsep+0.0in}
\setlength{\cellWidthd}{\columnwidth/5-2\tabcolsep-0.3in}
\setlength{\cellWidthe}{\columnwidth/5-2\tabcolsep-0.3in}
\scalebox{1}[1]{\begin{tabularx}{\columnwidth}{>{\PreserveBackslash\raggedright}m{\cellWidtha}>{\PreserveBackslash\centering}m{\cellWidthb}>{\PreserveBackslash\centering}m{\cellWidthc}>{\PreserveBackslash\centering}m{\cellWidthd}>{\PreserveBackslash\centering}m{\cellWidthe}}
\toprule
\textbf{Network Input} & \textbf{nuScenes} & \textbf{Rain and Night} & \textbf{Rain} & \textbf{Night} \\
\midrule

Lidar & {52.18}{\%} & {50.09}{\%}  & {43.94}{\%} & {63.56}{\%}\\
Radar & {17.43}{\%} &  {16.00}{\%}  & {16.42}{\%} &  {22.46}{\%}\\

Lidar, RGB & {49.96}{\%} & {46.59}{\%} & {42.72}{\%} & {61.66}{\%}\\
Lidar, Radar & {54.18}{\%} & {53.10}{\%}  & {47.51}{\%} & \textbf{{68.01}{\%}}\\
Lidar, RGB, Radar \textbf{(RVF-Net)} & \textbf{{54.86}{\%}} & \textbf{{53.12}{\%}} & \textbf{{48.32}{\%}} & {67.39}{\%}\\
Simulated Depth Cam & {48.02}{\%} &  {46.07}{\%} &  {39.07}{\%} & {57.33}{\%}\\
Simulated Depth Cam, Radar & {52.06}{\%} &  {48.31}{\%} & {41.65}{\%} & {61.04}{\%}\\
\bottomrule
\end{tabularx}
}
\end{specialtable}

The simulated depth camera approach does not increase the detection performance. The approach adds additional input data by depth-completing the lidar points. However, the informativeness in this data cannot compensate for the increased complexity introduced by its addition.

The absolute AP scores between the different columns of Table \ref{table:results_fusion} cannot be compared since the underlying data varies between the columns. The data source has the greatest influence for the performance of machine learning models. {All models have a significantly higher scores for the night scenes split than for the other splits. This is most likely due to the lower complexity of the night scenes present in the data set.}

The relative performance gain of different input data within each column shows a valid comparison of the fusion methods since they are trained and evaluated on the same data. The radar data fusion of the RVF-Net outperforms the lidar baseline by {5.1}{\%} on the nuScenes split, while it outperforms the baseline on the rain split by {10.0}{\%} and on the night split by {6.0}{\%}. The increased performance of the radar fusion is especially notable for the rain split where lidar and camera data quality is limited. The fusion of lidar and radar is also especially beneficial for night scenes, even though the lidar data quality should not be affected by these conditions.

\subsection{Ablation Studies}

This section evaluates additional training configurations of our proposed RVF network to measure the influence of the proposed training strategies. Table \ref{table:results_ablation} shows an overview of the results.

To study the effect of the introduced yaw loss, we measure the Average Orientation Error (AOE) as introduced by nuScenes. The novel loss reduces the orientation error by about \SI{40}{\percent} from an AOE of 0.5716 with the old loss to an AOE of 0.3468 for the RVF-Net. At the same time, our novel yaw loss increases the AP score of  RVF-Net by \SI{4.1}{percent}. Even though the orientation of the predicted bounding boxes does not directly impact the AP calculation, the simpler regression for the novel loss also implicitly  increases the performance for the additional regression targets.

Data augmentation has a significant positive impact on the AP score. 

Contrary to the literature results, the combined IoU and distance threshold decreases the network performance in comparison to a simple IoU threshold configuration. It is up to further studies to find the reason for this empirical finding.

We have performed additional experiments with 10 lidar sweeps as the input data.  While the sweep accumulation for static objects is not problematic since we compensate for ego-motion, the point clouds of moving objects are heavily blurred when considering 10 sweeps of data, as the motion of other vehicles cannot be compensated. Nonetheless, the detection performance increases slightly for the RVF-Net sensor input.

For a speed comparison, we have also started a training with non-sparse convolutions. However, this configuration could not be trained on our machine since the non-sparse network is too large and triggers an out-of-memory (OOM) error. 

\begin{specialtable}[H] 
\caption{AP scores for different training configurations on the validation data set.}
\label{table:results_ablation}

\setlength{\cellWidtha}{\columnwidth/2-2\tabcolsep+0.0in}
\setlength{\cellWidthb}{\columnwidth/2-2\tabcolsep+0.0in}
\scalebox{1}[1]{\begin{tabularx}{\columnwidth}{>{\PreserveBackslash\raggedright}m{\cellWidtha}>{\PreserveBackslash\centering}m{\cellWidthb}}

\toprule
\textbf{Network} & \textbf{nuScenes}\\
\midrule
RVF-Net & {54.86}{\%}\\
RVF-Net, simple yaw loss & {52.69}{\%}\\
RVF-Net, without augmentation & {50.68}{\%}\\
RVF-Net, IoU threshold only & {55.93}{\%}\\
RVF-Net, 10 sweeps &  {55.25}{\%}\\
% Lidar, 10 sweeps & {48.28}{\%}\\
RVF-Net, standard convolutions &  OOM error\\
\bottomrule
\end{tabularx}
}
\end{specialtable}

\subsection{Inference Time}
The inference time of the network for different input data configurations is shown in Table \ref{table:compute_times}. The GPU processing time per sample is averaged over all samples of the validation split. In comparison to the lidar baseline, the RVF-Net fusion increases the processing time only slightly. The different configurations are suitable for a real-time application with input data rates of up to \SI{20}{\hertz}. The processing time increases for the simulated depth camera input data configuration as the number of points is drastically increased by the depth completion.

\begin{specialtable}[H] 
\caption{Inference times of different sensor input configurations on the NVIDIA Titan Xp GPU.}
\label{table:compute_times}
\setlength{\cellWidtha}{\columnwidth/2-2\tabcolsep+0.0in}
\setlength{\cellWidthb}{\columnwidth/2-2\tabcolsep+0.0in}
\scalebox{1}[1]{\begin{tabularx}{\columnwidth}{>{\PreserveBackslash\raggedright}m{\cellWidtha}>{\PreserveBackslash\centering}m{\cellWidthb}}
\toprule
\textbf{Network Input} & \textbf{Inference Time}\\
\midrule
Lidar & \SI{0.042}{\second}\\
Radar & \SI{0.02}{\second}\\
Lidar, RGB& \SI{0.045}{\second}\\
Lidar, Radar & \SI{0.044}{\second}\\
RVF-Net & \SI{0.044}{\second}\\
Simulated Depth Cam, Radar & \SI{0.061}{\second}\\
RVF-Net, 10 sweeps  &  \SI{0.063}{\second}\\
\bottomrule
\end{tabularx}
}
\end{specialtable}

\subsection{Early Fusion vs. Late Fusion}
The effectiveness of the neural network early fusion approach is further evaluated against a late fusion scheme for the respective sensors. For the lidar, RGB, and radar input configurations are fused with an UKF and an Euclidean-distance-based matching algorithm to generate the final detection output. This late fusion output is compared against the early fusion RVF-Net and lidar detection results, which are individually tracked with the UKF to enable comparability. The late fusion tracks objects over consecutive time steps and requires temporal coherence for the processed samples, which is only given for the samples within a scene but not over the whole data set. Table \ref{table:results_late} shows the resulting AP score for 10 randomly sampled scenes to which the late fusion is applied. The sampling is done to lower the computational and implementation effort, and no manual scene selection in favor or against the fusion method was performed. The evaluation shows that the late fusion detection leads to a worse result than the early fusion. Notably, the tracked lidar detection outperforms the late fusion approach as well. As the radar-only detection accuracy is relatively poor and its measurement noise does not comply with the zero-mean assumption of the Kalman filter, a fusion of this data to the lidar data leads to worse results. In contrast to the early fusion where the radar features increased the detection score, the late fusion scheme processes the two input sources independently and the detection results cannot profit from the complementary features of the different sensors. In this paper, the UKF tracking serves as a fusion method to obtain detection metrics for the late fusion approach. It is important to note that for an application in autonomous driving, object detections need to be tracked independent of the data source, for example with a Kalman Filter, to create a continuous detection output. The evaluation of further tracking metrics will be performed in a future paper.

\begin{specialtable}[H] 
\caption{AP scores of the tracked sensor inputs. The early fusion RVF-Net outperforms the late fusion by a great margin.}
\label{table:results_late}
\setlength{\cellWidtha}{\columnwidth/2-2\tabcolsep+0.0in}
\setlength{\cellWidthb}{\columnwidth/2-2\tabcolsep+0.0in}
\scalebox{1}[1]{\begin{tabularx}{\columnwidth}{>{\PreserveBackslash\raggedright}m{\cellWidtha}>{\PreserveBackslash\centering}m{\cellWidthb}}
\toprule
\textbf{Network} & \textbf{nuScenes}\\
\midrule
Tracked Lidar & {40.01}{\%}\\
Tracked Late Fusion & {33.29}{\%}\\
Tracked Early Fusion (RVF-Net) & {47.09}{\%}\\
\bottomrule
\end{tabularx}
}
\end{specialtable}

%%%%%%%%%%%%%%%%%%%%%%%%%%%%%%%%%%%%%%%%%%
\section{Discussion}
\label{sec:discussion}

The RVF-Net early fusion approach proves its effectiveness by outperforming the lidar baseline by {5.1}{\%}. Additional measures have been taken to increase the overall detection score. Data augmentation especially increased the AP score for all networks. The novel loss, introduced in Section \ref{sec:loss}, improves both the AP score and notably the orientation error of the networks. Empirically, the additional classification loss mitigates the discontinuity problem in the yaw regression, even though classifications are discontinuous decisions on their own.

Furthermore, the paper shows that the early fusion approach is especially beneficial in inclement weather conditions. The radar features, while not being dense enough for an accurate object detection on their own, contribute positively to the detection result when processed with an additional sensor input. It is interesting to note that the addition of RGB data increases the performance of the lidar, radar, and camera fusion approach, while it does not increase the performance of the lidar and RGB fusion. We assume that the early fusion performs most reliably when more different input data and interdependencies are present. In addition to increasing robustness and enabling autonomous driving in inclement weather scenarios, we assume that early fusion schemes can be advantageous for special use cases such as mining applications, where dust oftentimes limits lidar and camera detection ranges.

When comparing our network to the official detection scores on the nuScenes data set, we have to take into account that our approach is evaluated on the validation split and not on the official test split. The hyperparameters of the network, however, were not optimized on the validation split, so that it serves as a valid test set. We assume that the complexity of the data in the frontal field of view does not differ significantly from the full 360 degree view. We therefore assume that the detection AP of our approach scales with the scores provided by other authors on the validation split. To benchmark our network on the test split, a 360 degree coverage of the input data would be needed. Though there are no conceptual obstacles in the way, we decided against the additional implementation overhead due to the general shortcomings of the radar data provided in the nuScenes data set~\cite{Scheiner.2020, Nobis.2021} and no expected new insights from the additional sensor coverage. The validation split suffices to evaluate the applicability of the proposed early fusion network.

On the validation split, our approach outperforms several single sensor or fusion object detection algorithms. For example, the CenterFusion approach~\cite{Nabati.10.11.2020}, which achieves \SI{48.4}{\percent} AP for the car class on the nuScenes validation split. In the literature, only Wang~\cite{Wang.1019202011132020} fuses all three sensor modalities. Our fusion approach surpasses their score of \SI{45}{\percent} AP on the validation split and \SI{48}{\percent} AP on the test split. 

On the other hand, further object detection methods, such as the leading lidar-only method CenterPoint~\cite{Yin.19.06.2020}, outperform even our best network in the ablation studies by a great margin. The two stage network uses center points to match detection candidates and performs an additional bounding box refinement to achieve an AP score of {87}{\%} on the \mbox{test split.}

\begin{figure}[H]
    
    \includegraphics{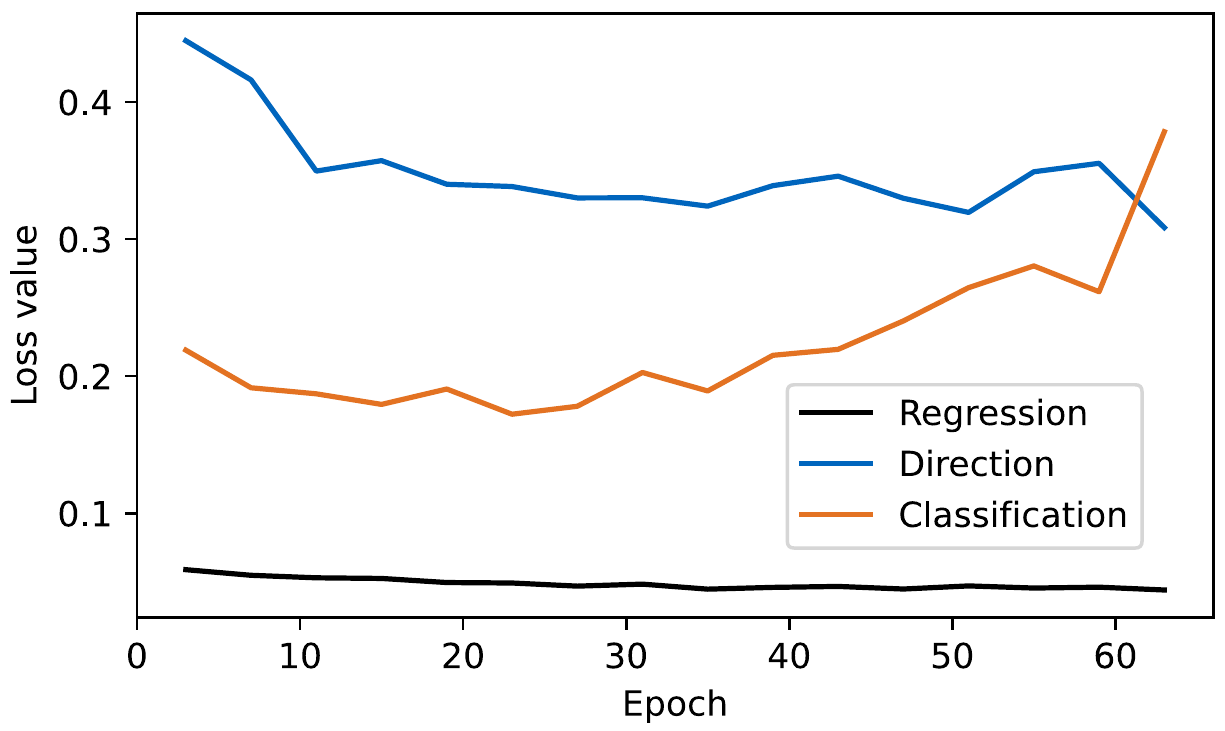}
    \caption{Loss values of the RVF-Net. The classification loss starts to overfit around epoch 30 while regression and direction loss continue to converge.}
    \label{fig:loss_overfit}
\end{figure}

When analyzing the errors in our predictions, we see that the regressed parameters of the predicted bounding boxes are not as precise as the ones of state-of-the-art networks. The validation loss curves for our network are shown in Figure \ref{fig:loss_overfit}. The classification loss overfits before the regression loss converges. Further studies need to be performed in order to further balance the losses. One approach could be to first only train the regression and direction loss. The classification loss is then trained in a second stage. Additionally, further experiments will be performed to fine tune the anchor matching thresholds to the data set to get a better detection result. The tuning of this outer optimization loop requires access to extensive GPU power to find optimal hyperparameters. For future work, we expect the hyperparameters to influence the absolute detection accuracy greatly as simple strategies such as data augmentation could already improve the overall performance. The focus of this work lies in the evaluation of different data fusion inputs relative to a potent baseline network. For this evaluation, we showed a vast amount of evidence to motivate our fusion scheme and network parameterization.

The simulated depth camera did not provide a better detection result than the lidar-only detection. This {and the late fusion approach} show that a simple fusion assumption in the manner of "more sensor data, better detection result" does not hold true. The complexity introduced by the additional data decreased the overall detection result. The decision for an early fusion system is therefore dependent on the sensors and the data quality available in the sensors. For all investigated sub data sets, we found that early fusion of radar and lidar data is beneficial for the overall detection result.
Interestingly, the usage of 10 lidar sweeps increased the detection performance of the fusion network over the proposed baseline. This result occurred despite the fact that the accumulated lidar data leads to blurry contours for moving objects in the input data. This is especially disadvantageous for objects moving at a high absolute speed. For practical applications, we therefore use only three sweeps in our network, as the positions of moving objects are of special interest for autonomous driving. The established metrics for object detection do not account for the importance of surrounding objects. We assume that the network trained with 10 sweeps performs worse in practice, despite its higher AP score. Further research needs to be performed to establish a detection metric tailored for autonomous driving applications. 

The sensors used in the data set do not record the data synchronously. This creates an additional ambiguity in the input data between the position information inferred from the lidar and from the radar data. The network training should compensate for this effect partially; however, we expect the precision of the fusion to increase when synchronized sensors \mbox{are available.}

This paper focuses on an approach for object detection. Tracking/prediction is applied as a late fusion scheme or as a subsequent processing step to the early fusion. In contrast, LiRaNet~\cite{Shah.02.10.2020} performs a combined detection and prediction of objects from the sensor data. We argue that condensed scene information, such as object and lane positions, traffic rules, etc., are more suitable for the prediction task in practice. A decoupled detection, tracking, and prediction pipeline increases the interpretability of all modules to facilitate validation for real-world application in autonomous driving.

%%%%%%%%%%%%%%%%%%%%%%%%%%%%%%%%%%%%%%%%%%
\section{Conclusions and Outlook}
\label{sec:conclusions}
In this paper, we have developed an early fusion network for lidar, camera, and radar data for 3D object detection. This early fusion network outperforms both the lidar baseline and the late fusion of lidar, camera, and radar data on a public autonomous driving data set. In addition, we integrated a novel loss for the yaw angle regression to mitigate the effect of the discontinuity of a simple yaw regression target. We provide a discussion about the advantages and disadvantages of the proposed network architecture. Future steps include the amplification of the fusion scheme to a full 360 degree view and the optimization of hyperparameters to balance the losses for further convergence of the regression losses.

We have made the code for the proposed network architectures and the interface to the nuScenes data set available to the public. The repository can be found online at \href{https://github.com/TUMFTM/RadarVoxelFusionNet}{https://github.com/TUMFTM/RadarVoxelFusionNet}.

%%%%%%%%%%%%%%%%%%%%%%%%%%%%%%%%%%%%%%%%%%
\vspace{6pt}

%%%%%%%%%%%%%%%%%%%%%%%%%%%%%%%%%%%%%%%%%%
\authorcontributions{F.N., as the first author, initiated the idea of this paper and contributed essentially to its concept and content. Conceptualization, F.N.; methodology, F.N. and E.S.; software, E.S., F.N. and P.K.; data curation, E.S. and F.N.; writing---original draft preparation, F.N.; writing---review and editing, E.S., P.K., J.B. and M.L.; visualization, F.N. and E.S.; project administration, J.B. and  M.L. All authors have read and agreed to the published version of the manuscript.}

%%%%%%%%%%%%%%%%%%%%%%%%%%%%%%%%%%%%%%%%%%
\funding{We express gratitude to Continental Engineering Services for funding for the underlying research project.}

%%%%%%%%%%%%%%%%%%%%%%%%%%%%%%%%%%%%%%%%%%
%\acknowledgments{In this section you can acknowledge any support given which is not covered by the author contribution or funding sections. This may include administrative and technical support, or donations in kind (e.g., materials used for experiments).}

%%%%%%%%%%%%%%%%%%%%%%%%%%%%%%%%%%%%%%%%%%
\conflictsofinterest{The authors declare no conflict of interest.} 
%%%%%%%%%%%%%%%%%%%%%%%%%%%%%%%%%%%%%%%%%%

\end{paracol}

\reftitle{References}


\externalbibliography{yes}

%%%%%%%%%%%%%%%%%%%%%%%%%%%%%%%%%%%%%%%%%%
%% for journal Sci
%\reviewreports{\\
%Reviewer 1 comments and authors’ response\\
%Reviewer 2 comments and authors’ response\\
%Reviewer 3 comments and authors’ response
%}
%%%%%%%%%%%%%%%%%%%%%%%%%%%%%%%%%%%%%%%%%%
\end{document}